# An Intelligent Pixel Replication Technique by Binary Decomposition for Digital Image Zooming

Kaeser Md. Sabrin
Dept. of Computer Science & Engineering
University of Dhaka
Dhaka, Bangladesh
kmsabrin@gmail.com

Md. Haider Ali
Dept. of Computer Science & Engineering
University of Dhaka
Dhaka, Bangladesh
haider@univdhaka.edu

*Abstract*— Image zooming is the process of enlarging the spatial resolution of a given digital image. We present a novel technique that intelligently modifies the classical pixel replication method for zooming. Our method decomposes a given image into layer of binary images, interpolates them by magnifying the binary patterns preserving their geometric shape and finally aggregates them all to obtain the zoomed image. Although the quality of our zoomed images is much higher than that of nearest neighbor and bilinear interpolation and comparable with bicubic interpolation, the running time of our technique is extremely fast like nearest neighbor interpolation and much faster than bilinear and bicubic interpolation.

## I. Introduction

In this modern era of high performance computing machineries, imaging application are becoming ubiquitous in our everyday life. Image processing techniques that magnifies a digital image is routine process encountered in diverse applications like medical imaging, astronomy and satellite imaging, law enforcement, military defense, industrial operation, archaeological research, high definition visualization, large size printing, digital and mobile phone cameras and many more. Whenever the user wishes to gain a closer and better look in the above mentioned applications, images will be required to be zoomed in and out and reproduced to higher resolution from lower resolution.

The problem of image magnification has been pursued in a variety of ways in the literature. Most current imaging software packages as well as software included in digital cameras implement the conventional linear magnification algorithms. Among them nearest neighbor interpolation, bilinear interpolation and bicubic interpolation methods are most notable [1-3]. Nearest neighbor interpolation method is a technique of blind pixel replication, which is simple to implement by replicating the original pixels. This method is usually susceptible to the undesirable defect of blocking effects. Bilinear and bicubic interpolation employ first-order spline and second-order spline model, respectively to achieve a more pleasing outcome, although bilinear interpolation tends to blur the zoomed image. Among the above mentioned techniques nearest neighbor is the fastest one, whereas both bilinear and bicubic interpolation method has considerable computational complexities preventing them from being utilized in real time systems like HDTV or web video streaming and hand-held devices with low processing capabilities [4].

To this end, we have redesigned the simple pixel replication method from a completely new perspective and proposed a novel technique based on binary decomposition. Our algorithm's computational complexity is extremely low, like that of nearest neighbor interpolation, whereas the quality of the magnified image clearly outperforms nearest neighbor and bilinear interpolation zooming and comparable with bicubic interpolation zooming. This enables our technique to be employed in high speed real-time systems as well as low-capability mobile devices.

The paper is organized as follows: our binary decomposition based intelligent pixel replication method is introduced in Section II. Section III is devoted to experimental results, followed by some conclusions in Section IV.

## II. Intelligent Pixel Replication by Binary Decomposition

In our proposed method we have introduced an intelligent modification in the classical pixel replication zooming technique by using the concept of binary decomposition. We begin by describing our magnification framework. Given an input image L with resolution $w \times h$, we want to infer the zoomed image H having resolution $(2w - 1) \times (2h - 1)$ to achieve a 2 times magnification as shown in Figure 1. Note that we can repeatedly apply the same procedure to achieve 4, 8, 16, … $2^n$ time's magnification.

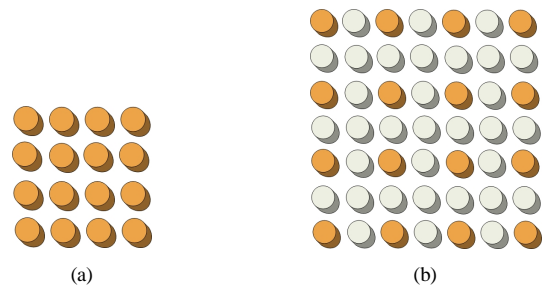

Figure 1. (a) Input Image (b) 2× Magnified Image

As can be seen from Figure 1(b), due to zooming there will be newly created pixels (colored white) with undefined intensity. The purpose of a zooming algorithm will be to infer values for those undetermined pixels from nearby known pixels in a way that is fast and accurate.

*A. Motivations*

Referring to Figure 2, we start with a given *2 × 2* low resolution (LR) image patch with pixel values {a, b, c, d} and we would like to infer the *3 × 3* high resolution (HR) image patch with pixels {a, b, c, d} positioned as shown and **5** (five) newly created pixels. Now most existing interpolation/zooming techniques [5-9] use the exact values of {a, b, c, d} in their interpolation which we have found causes the following intricacies.

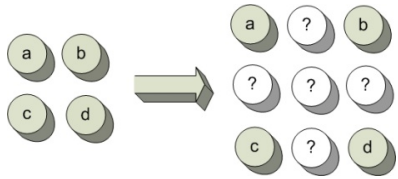

Figure 2. Image Zooming Scenario in Existing Methods

1. Each of the {a, b, c, d} pixels can have values ranging from [0-255], giving us a possible $256^4 = 4294967296$ possible patterns that can occur in a LR patch. Encompassing these large variations within some general rule is very difficult.
2. When exact values of {a, b, c, d} are used, multiple heuristics are required for determining whether one pixel value is sufficiently close to another and often variations in image illumination can cause instability for the chosen heuristics.
3. For sharp magnification, some sort of edge detection method is usually employed which introduces large computational complexity.
4. Determined new pixel values tend to be statistical in nature (mean, median, weighted average etc.) that usually causes blurriness.

*B. Proposed Methods*

Theoretically our method has the following three stages as depicted in Figure 3.

**1. Binary Decomposition:** We threshold input image L of size $W \times H$ using threshold parameter T with values [0-255] producing 256 binary images using the following equation.

$$L_i = \begin{cases} 1 & \text{if } T_i < L(x,y) \\ 0 & \text{otherwise} \end{cases}$$

where, $i = 0 \to 255$, $x = 0 \to W-1$, $y = 0 \to H-1$ (1)

**2. Binary Interpolation:** We magnify each of the 256 binary images, $L_i$ into the size of magnified image H using a special binary pattern interpolation method to be explained shortly.

$$L_i \to Binary\,Enlarging \to H_i, \quad i = 0 \to 255 \quad (2)$$

**3. Aggregation:** Finally we sum up each of the 256 magnified binary images $H_i$ to get the final zoomed high resolution image, H.

$$\sum_{i=0}^{255} H_i \to H \quad (3)$$

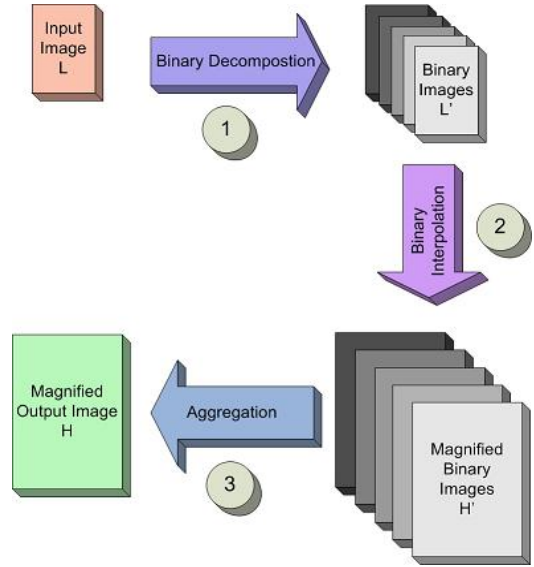

Figure 3. Steps of the Proposed Intelligent Pixel Replication Technique

We now describe in detail binary interpolation method. Using binary decomposition we have eliminated exact pixel values of large magnitude and in each of our binary images only possible pixel values are 0 and 1. So now, a *2 × 2* LR image patch can have only $2^4 = 16$ possible patterns instead of $256^4$ patterns. We interpolate each of these 16 LR *2 × 2* sized binary patches into *3 × 3* sized HR binary patches in a way that preserves the inherent geometrical shapes they represent. This process is depicted in Figure 4 and explained in the figure caption.

The process of binary interpolation has the following desirable properties,
1. It preserves the sharp edges and boundaries in an elegant way without producing any blocking effect.
2. The smooth regions are also preserved without blurring the whole image.
3. The method does not produce any noise, artifacts or variations that was not present in the original image.

Finally when all the 256 binary LR images have been interpolated into corresponding binary HR images, we simply aggregate (mathematical summation) all the HR images to construct the final zoomed HR image. The validity of the final summation is evident from how we initially decomposed our input image.

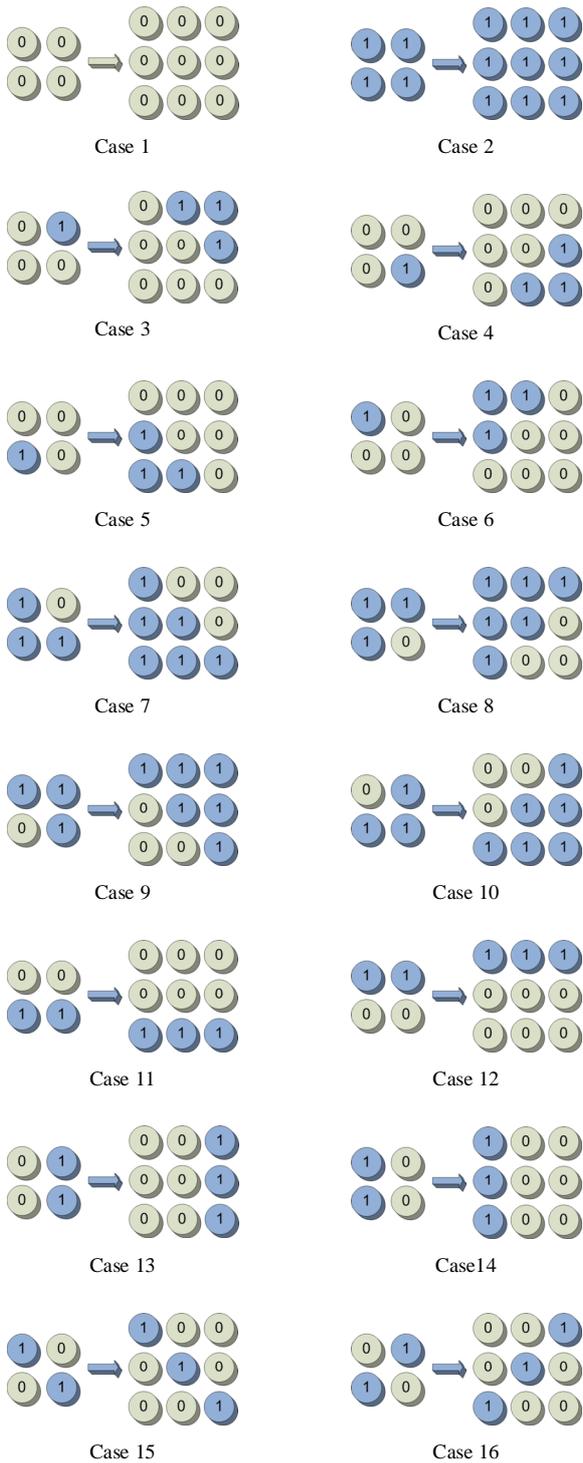

Figure 4. Case 1 and 2 represent a rectangular shaped binary pattern enlarged keeping the filled shape. Case 3-10 represent a triangular shape situated at various corners that are expanded preserving the triangle shape. Case 11-14 represent horizontally or vertically similar shapes that are expanded maintaining the similarity stretch. Case 15 and 16 represent diagonally separable 3 segment shapes which are expanded maintaining the separation and segmentation.

If we try to employ straightforwardly the techniques described so far, it will incur a huge computational complexity. This is obvious since we are interpolating 256 images instead of a single image. Interestingly, the algorithm proposed so far has a nice closed form mathematical representation. That will enable us to achieve the effect of all separate 256 binary interpolations and final aggregation through a single operation only. We shall now derive the stated mathematical formulation.

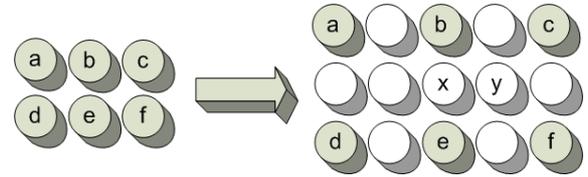

Figure 5. Illustration required for Deriving a Mathematical Closed Form

Referring to Figure 5, {a, b, c, d, e, f} are original pixel values coming from input LR image. We seek to find the values of pixels {x, y} (all other pixels will have symmetrical calculations). We have the following claims,

(1) x = (b∩e) and (2) y = (b∩ f) ∪ (c∩e).

Here the operation (p∩q) returns the minimum value between p and q and the operation (p∪q) returns the maximum value between p and q.

*Proof:*

According to our claim (b∩e) is the number of binary HR images having values 1 at pixel x. As can be seen from Figure 5, x is effected by rectangular, triangular and vertical line shaped binary patterns as explained in our algorithm. The rectangular pattern (case 1-2) existed (a∩b∩d∩e) and (b∩c∩e∩f) times, the triangular pattern (case 3-10) existed (a∩b∩e), (b∩d∩e), (b∩c∩e) and (b∩f∩e) times and the vertical line pattern (case 13-14) existed (b∩e) times. x does not involve any diagonal patterns.

Now, x = (a∩b∩d∩e) ∪ (b∩c∩e∩f) ∪ (a∩b∩e) ∪ (b∩d∩e) ∪
(b∩c∩e) ∪ (b∩f∩e) ∪ (b∩e)

= (a∩b∩e) ∪ (b∩d∩e) ∪ (b∩c∩e) ∪ (b∩f∩e) ∪ (b∩e)

= (b∩e)   [repeatedly applying (x ∩ y) ∪ y = y]

In a similar manner, we can see for y, the rectangular pattern existed (b∩c∩e∩f) times, the triangular pattern existed (b∩c∩e), (b∩e∩f), (b∩c∩f) and (c∩e∩f) times, the diagonal pattern (case 15-16) existed (b∩f) and (c∩e) times. y does not involve any horizontal/vertical line shaped pattern.

Now, y = (b∩c∩e∩f) ∪ (b∩c∩e) ∪ (b∩e∩f) ∪ (b∩c∩f) ∪
(c∩e∩f) ∪ (b∩f) ∪ (c∩e)

= (b∩c∩e) ∪ (b∩e∩f) ∪ (b∩c∩f) ∪ (c∩e∩f) ∪ (b∩f) ∪
(c∩e)

= (b∩f) ∪ (c∩e)  [repeatedly applying (x ∩ y) ∪ y = y]
∎

We now give our final zooming algorithm in Figure 6(a) which refers to Figure 6(b).

*Input: Low resolution image L*
*Output: High resolution image H*
1. *for all undefined pixels in the enlarged image H*
2.     *if the pixel is an undefined vertical pixel, x*
3.         *then, x = Min( a, c )*
4.     *if the pixel is an undefined horizontal pixel, y*
5.         *then, y = Min( a, b )*
6.     *if the pixel is an undefined central pixel, z*
7.         *then, z = Max( Min( a, d ), Min( b, c ) )*
8. *end for*

(a)

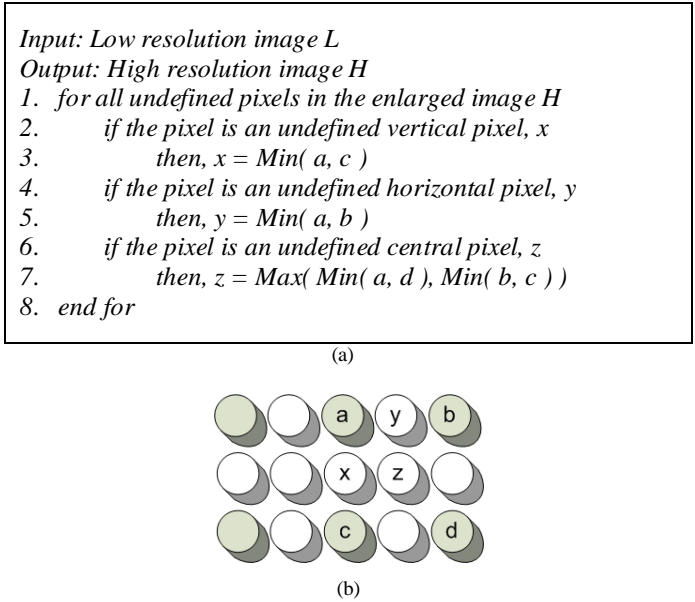

(b)

Figure 6. (a) Intelligent Pixel Replication Algorithm (b) Reference Image for Illustration

For color image zooming we would apply our algorithm separately on R, G and B channel and combine them at the end. Our algorithm runs in linear time $O(n)$ where n is the output image size (in pixels) since we require $O(1)$ computation time per new pixel value determination. The space complexity is also $O(n)$ because only as much memory as the size of output image is required. Note that for a magnified image of size n, $O(n)$ is the lowest possible complexity achievable. Also note that since the algorithm requires only a single pass and localized pixel operations, it readily lends itself to complete parallelization at each pixel simultaneously. With no floating point calculation involved, our method is, to our knowledge amongst the fastest image zooming techniques with acceptable quality output.

III. EXPERIMENTAL RESULTS

We have performed several experiments to assess the performance of our proposed method and compared it with well established zooming methods like nearest neighbor, bilinear and bicubic interpolation. We shall present our findings in the next three subsections.

*A. Visual Quality*

From visual inspection our method produces better images than nearest neighbor and bilinear method and is quite close to bicubic method. Our test set consisted of a number of both grey level and color images. We have also used a set of highly textured image and text images to demonstrate the capability of our method in suppressing the blocking effect and maintaining sharp edges. In all our images, we had a ground truth image of size $256 \times 256$. We decimated that image into the size of $128 \times 128$ which we then zoomed using different methods. Figure 7, 8 and 9 shows the zoomed images of the nearest neighbor, bilinear and bicubic methods compared with our method. Quality of zooming in our method is visible in the digits and letters of the 'Calculator' and 'Vision' image and sharp object borders in the 'Stones' image. More experimental zoomed images, in their actual sizes, will be provided for visual comparison if requested.

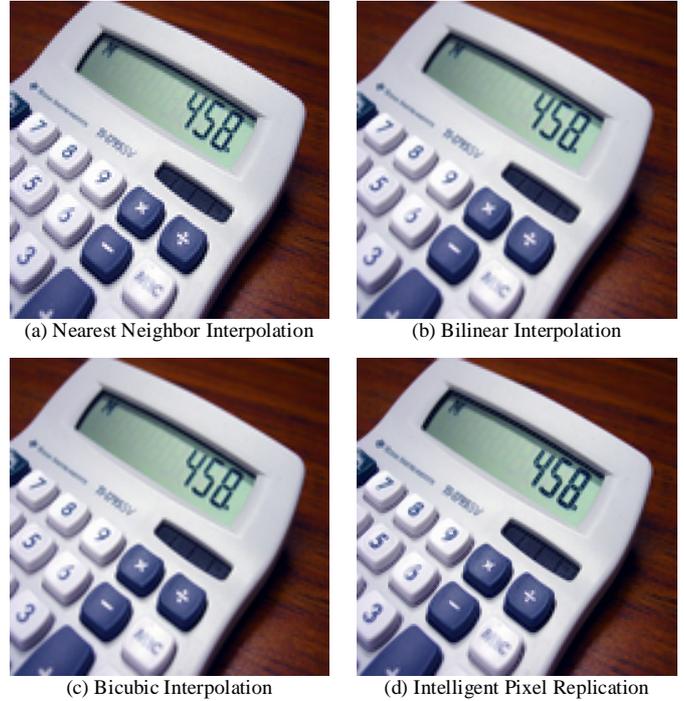

(a) Nearest Neighbor Interpolation      (b) Bilinear Interpolation

(c) Bicubic Interpolation      (d) Intelligent Pixel Replication

Figure 7. Comparison of Zooming Methods on 2× Zoomed 'Calculator' Image

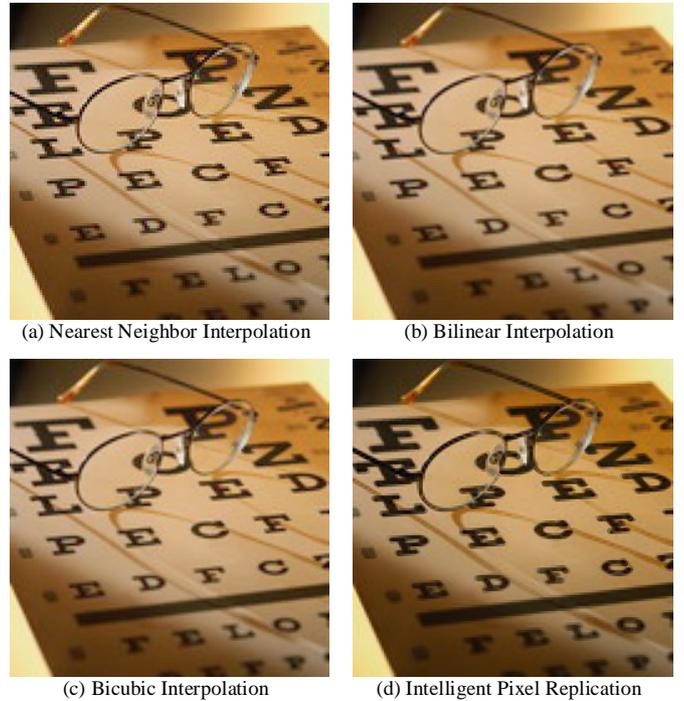

(a) Nearest Neighbor Interpolation      (b) Bilinear Interpolation

(c) Bicubic Interpolation      (d) Intelligent Pixel Replication

Figure 8. Comparison of Zooming Methods on 2× Zoomed 'Vision' Image

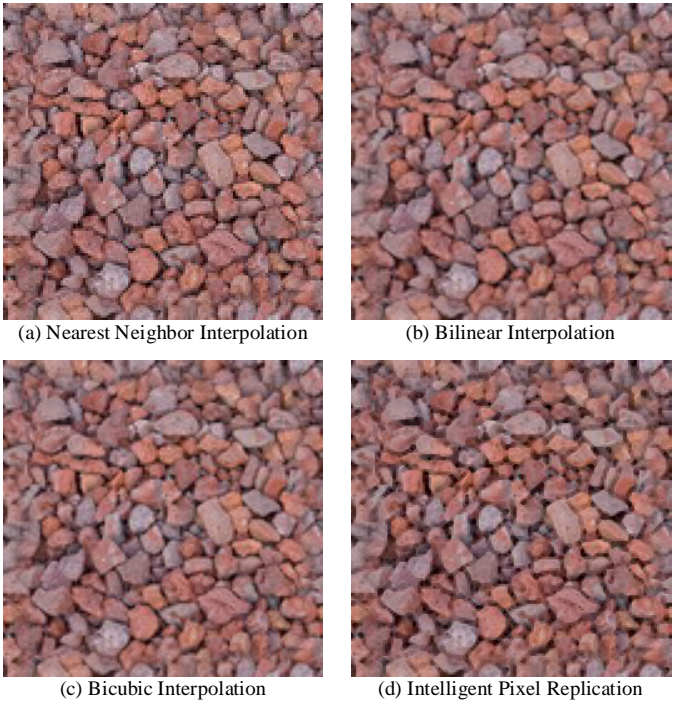

(a) Nearest Neighbor Interpolation  (b) Bilinear Interpolation
(c) Bicubic Interpolation  (d) Intelligent Pixel Replication

Figure 9. Comparison of Zooming Methods on 2× Zoomed 'Stones' Image

### B. Quantitative Analysis

We have used the classical PSNR measures between the original picture and the reconstructed picture to assess the numerical quality of zooming [2, 10]. The PSNR is defined in the following equations where $I$ is the original image and $I'$ is zoomed image.

$$MSE = \frac{1}{m \times n} \sum_{x=0}^{m-1} \sum_{y=0}^{n-1} (I'(x,y) - I(x,y))^2 \quad (4)$$

$$PSNR = 20 \times \log_{10} \frac{255^2}{MSE} \quad (5)$$

Table 1 gives the PSNR values of different methods for the set of images presented in this paper. We can see that our method consistently outperforms both pixel replication and bilinear method and compares closely with bicubic method.

TABLE 1. PSNR VALUES OF VARIOUS ZOOMING METHODS ON TEST IMAGES

| Zooming Technique | PSNR (dB) | | |
|---|---|---|---|
| | *Calculator* | *Vision* | *Stones* |
| **Nearest Neighbor Interpolation** | 26.63 | 32.51 | 33.22 |
| **Bilinear Interpolation** | 28.09 | 34.66 | 34.26 |
| **Bicubic Interpolation** | 29.64 | 36.47 | 37.71 |
| **Intelligent Pixel Replication** | 28.98 | 35.11 | 36.43 |

### C. Computational Complexity

One of the major efficiency of our proposed method comes from its extremely low computational cost. Table 2 reports the number of operations per pixel for a straight forward implementation of nearest neighbour, bilinear and bicubic interpolation methods compared with our method [11]. The fact that our method produces good quality zooming with such little computational burden indicates its suitability in real-time and embedded applications.

TABLE 2. NUMBER OF OPERATIONS PER PIXEL FOR DIFFERENT ZOOMING METHONDS

| Zooming Method | Operations | |
|---|---|---|
| | *Addition* | *Multiplication* |
| **Nearest Neighbor Interpolation** | 2 | 0 |
| **Bilinear Interpolation** | 16 | 18 |
| **Bicubic Interpolation** | 22 | 29 |
| **Intelligent Pixel Replication** | 0 | 0 |

Comparison operations are ignored here

### IV. CONCLUSION

In this paper we have proposed a technique for zooming digital images. We transformed the blind pixel replication process into an intelligent zooming algorithm by introducing the concept of shape preserving binary decomposed interpolation. Complexity analysis suggests that ours is amongst the fastest zooming techniques available. Experimental results shows that even with such simplistic nature of the algorithm, it is capable of producing good quality zooming comparable to computationally expensive bilinear and bicubic interpolation methods.